\documentclass[letterpaper, conference]{ieeeconf}
\pdfoutput = 1

\usepackage{amsmath,amssymb}
\usepackage{todonotes}
\usepackage{color}
\usepackage{hyperref}
\allowdisplaybreaks
\usepackage{float}
\usepackage[noadjust]{cite}
\usepackage{dsfont}
\usepackage{amsthm}
\usepackage{algorithmic}
\usepackage{tikz}
\usepackage{graphicx}
\usepackage[ruled,vlined,shortend, linesnumbered]{algorithm2e} 

\newtheorem{remark}{\bf Remark}

\newcommand{\hide}[1]{}

\newcommand{\bit}{\begin{compactitem}}
\newcommand{\eit}{\end{compactitem}}
\newcommand{\ben}{\begin{compactenum}}
\newcommand{\een}{\end{compactenum}}

\renewcommand{\S}{\mathcal{S}}

\newcommand{\V}{\mathcal{V}}
\newcommand{\E}{\mathcal{E}}
\newcommand{\N}{\mathcal{N}}

\usepackage{xcolor}

\begin{document}

\title{Anomaly Detection in Power Grids via Context-Agnostic Learning}

\author{SangWoo Park and Amritanshu Pandey
\thanks{SangWoo Park is an Assistant Professor in the Department of Mechanical and Industrial Engineering, New Jersey Institute of Technology,  Newark, NJ, USA
        {\tt\small sangwoo.park@njit.edu}}%
\thanks{Amritanshu Pandey is an Assistant Professor in the Department of Electrical Engineering, University of Vermont, Burlington, VT, USA
        {\tt\small amritanshu.pandey@uvm.edu}}%
}

\maketitle

\begin{abstract}
An important tool grid operators use to safeguard against failures, whether naturally occurring or malicious, involves detecting anomalies in the power system SCADA data. In this paper, we aim to solve a real-time anomaly detection problem. Given time-series measurement values coming from a fixed set of sensors on the grid, can we identify anomalies in the network topology or measurement data? Existing methods, primarily optimization-based, mostly use only a single snapshot of the measurement values and do not scale well with the network size. Recent data-driven ML techniques have shown promise by using a combination of current and historical data for anomaly detection but generally do not consider physical attributes like the impact of topology or load/generation changes on sensor measurements and thus cannot accommodate regular context-variability in the historical data. To address this gap, we propose a novel context-aware anomaly detection algorithm, GridCAL, that considers the effect of regular topology and load/generation changes. This algorithm converts the real-time power flow measurements to context-agnostic values, which allows us to analyze measurement coming from different grid contexts in an aggregate fashion, enabling us to derive a unified statistical model that becomes the basis of anomaly detection. Through numerical simulations on networks up to 2383 nodes, we show that our approach is accurate, outperforming state-of-the-art approaches, and is computationally efficient.
\end{abstract}

\section{Introduction} \label{sec:intro}
The outputs from AC state-estimation (SE) and network topology processor (NTP) are critical for i) monitoring the critical grid states, 
ii) evaluating the resilience of the grid operating state via what-if analysis (real-time contingency analysis), and 
iii) delivering the network model for downstream market mechanisms. 
Subsequently, North American Reliability Corporation (NERC) considers the malfunction or loss of SE or NTP as a severe threat to the bulk power system (BPS) reliability, as it impacts the required real-time grid assessment every 30 minutes \cite{nerc_rta}. 

While the criticality of NTP+SE is well established, at present, the NTP+SE solution quality is highly sensitive to the accuracy of the switch status and measurement data. 
Failure and malfunction of NTP+SE is a common occurrence \cite{nerc_ll1, nerc_ll2, nerc_ll3} and \textbf{anomalous} measurements and switch status data are one of the primary causes of these failures. 
For instance, in NERC lessons learned \cite{nerc_ll3}, a grid transmission operator reported via \textit{post-event analysis} that the SE diverged due to incorrectly telemetered MW/MVAR.

Research output from both industry and academia have proposed many techniques to identify and isolate anomalous network topology and measurements \cite{statistical_anomaly_detection_heydari, Dynwatch_2021, schweppe_bad_data}. 
The underlying philosophy of these methods can be broadly classified into: 
i) traditional mechanistic methods that apply grid physics-based rules to identify anomalous grid topology and measurements \cite{schweppe_bad_data, takagi1982_fault_location, niemira2013malicious, liao2007unsynchronized, peterson1988multiple} and 
ii) emerging data-driven methods that apply data mining advancements to detect anomalous grid topology and measurements \cite{heydari_mrf, Dynwatch_2021}. 
Many successful methods \cite{Dynwatch_2021, hooi2018gridwatch} combine the features of both traditional mechanistic and emerging data-mining techniques. While earlier anomaly detection methods were generally designed for non-interacting bad-data, newer approaches are designed for interacting bad-data such as those observed during coordinated attacks on grid infrastructure.

Much of traditional mechanistic approaches are based on the statistical property of the residual and optimization outputs from the underlying  AC-state estimation (ACSE) \cite{schweppe_bad_data, niemira2013malicious, lourencco2004bayesian, lourencco2006topology, singh2005topology}. 
NTP inputs the grid circuit breaker/switching status data and converts it into a bus-branch model. 
ACSE operates on the bus-branch model and the measurement data to output the residual information.
Under anomalous switch status or measurement data, the residual output will deviate from its chi-square distribution threshold and be flagged appropriately. The work~\cite{clements1998topology} used normalized Lagrange multipliers of the least-squares state estimation problem, and despite being a heuristic method, has been shown to be effective in some cases. Later studies, such as~\cite{lin2016computationally}, improved on this approach.

Other advanced mechanistic approaches apply robust (W)LAV-based estimators \cite{LAVSE-PMU-abur, li2021wlav, NLAV_2021}.
The methodology may fail when anomalies are coordinated (i.e., there are interacting bad data) and if switch status and measurement anomalies are sufficiently large to cause ACSE divergence.
Convexification of algorithms has been proposed to improve problem convergence \cite{convex_se, convexTESE-SDP-weng, convexSE-LAV-Li}, but these remain highly sensitive to switch-status anomalies and infeasible-AC solutions (in most cases).

Alternatively, time-series measurement data can be used for anomaly detection by analyzing and modeling the underlying statistical behavior \cite{hamilton1994time}.
Traditional data mining techniques can be directly applied to learn the statistical model of the grid states, and anomalies can be flagged if certain measurement and switch status data deviate from its statistical norm. However, measurement data depend on the ever-changing grid context (network topology and loads) and learning a statistical model that accurately captures the conditional distributions is extremely difficult. Without considering the underlying grid context and the grid physics, false positives are common \cite{LOFvariants} and the usefulness of the techniques is low. 
\textit{Context-aware} techniques that consider the actual physics of the grid \cite{Dynwatch_2021, song2017powercast} can overcome the drawbacks of both mechanistic and purely data-driven approaches.

\cite{hooi2018gridwatch}, for instance, considers the physical network of the power grid by embedding the edge pairing of the transmission lines and transformers into the method. 
However, this method \cite{hooi2018gridwatch} does not account for the dynamic nature of the power grid network and corresponding topology adjustments. 
\cite{Dynwatch_2021} builds on \cite{hooi2018gridwatch} and considers the dynamic nature of the power grid graph. 
It introduces a concept of \textit{graph distance}, which uses weights to quantify the \textit{similarness} of prior grid topologies to current grid topology. 
However, this approach assumes load and generation time-series data to be stationary within each analyzed period and does not account for adjustments beyond white noise. The methodology analyzes the distribution of flow change over consecutive time steps, which works well when load/generation is static but fails when load/generation variation is steep.
We also find (see Results) that \cite{Dynwatch_2021} algorithm is susceptible to high false-positive rates (Type 1). 
While the proposed work closely builds on these works, many other methods exist for grid anomaly detection. 
But in prior works, \cite{hooi2018gridwatch} and \cite{Dynwatch_2021}  have outperformed these generic methods \cite{parzen1962estimation}, \cite{LOFvariants}, and \cite{hamilton1994time}. Therefore, we focus on improving these state-of-the-art methods \cite{hooi2018gridwatch, Dynwatch_2021}.

This paper presents a \textit{context-aware} anomaly detection algorithm based on unsupervised learning, called GridCAL (Grid Context-Agnostic Learning). Instead of developing a statistical model based on raw measurement values, we propose \textit{a mapping} that converts the real-time line and transformer flow measurement data into \textit{context-agnostic} flow values, which then allows us to analyze data coming from the dynamic context in an aggregate fashion. The transformed data is called \textit{context-agnostic} because it is independent of grid topology and load/generation levels.

\section{Methodology} \label{sec:methods}

We are given a dynamic graph (grid) $\mathcal{G}_{\tau}=(\V,\E(\tau))$ at each time tick $\tau$, where $\V$ denotes the set of nodes (active grid buses), and $\E(\tau)$ denotes the set of edges (active grid branches). Note that the change over time is only in the edge-set, which is sufficient enough to also capture the changes in the node-set. This sequence of changing graph data is obtained from the network topology processor and does not account for topology errors due to erroneous status data caused by miscommunication, operator entry errors, cyber-attacks, etc. Therefore, at any time $\tau$, the given dynamic graph $\mathcal{G}(\tau)$ can be considered a `presumed topology'.  

\begin{figure}[t]
	\centering
	\includegraphics[width=0.9\linewidth, trim={0 0.8cm 0 0},clip]{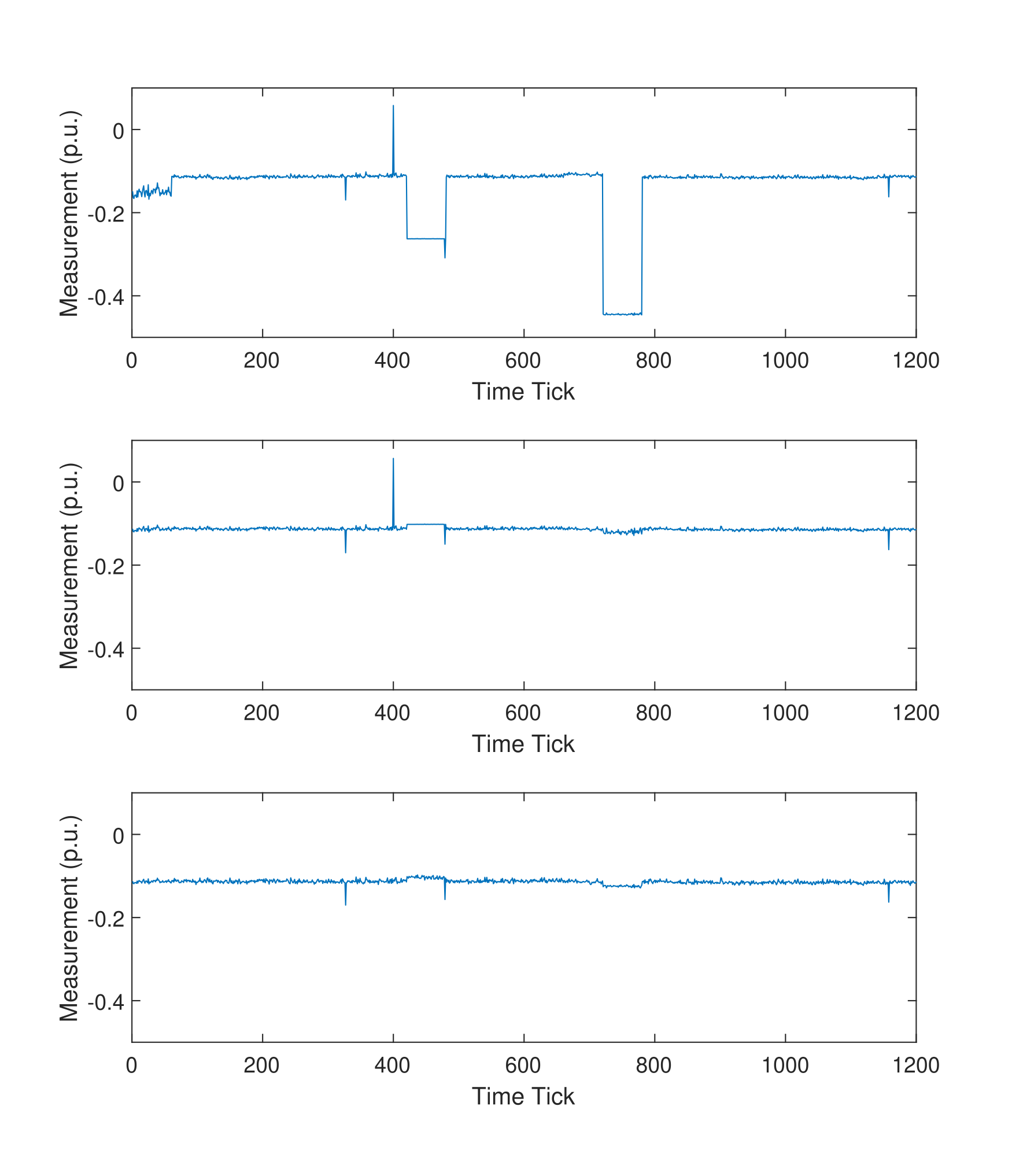}
	\caption[]{Flow signal before any processing (top),  after only inverse projection (middle), and after injection-correction and inverse projection (bottom) on line 89 (\#10 among current edgeset).}
	\label{fig:flow_comparison_line89_loadcorrection}
\end{figure}

Also, we have a fixed set of sensors $\S \subseteq \V$. Each sensor installed on node $i$ can obtain PMU or RTU measurements at each time $\tau$.  For each sensor on node $i$, we obtain the power flows on all lines and transformers adjacent to node $i$ since, as observed in \cite{hooi2018gridwatch}, using power (rather than current) provides better anomaly detection in practice. For any PMU bus $i$ and edge $e \in \N_i$, define the power w.r.t. $i$ along edge $e$ as $s_{ie}(\tau) = V_i(\tau) \cdot I_e(\tau)^*$, where $^*$ is the complex conjugate. The set of edges where the power flows are measured by sensors $\S$ is denoted by $\E_{\S}$. Without loss of generality, we assume that topology errors happen on lines that are not measured since detecting line status changes on measured lines is relatively straightforward.  

If the observed measurements do not contain any anomaly, they should be consistent with the `presumed topology' as long as there are no topology errors. This work uses the output from topology estimation (NTP) in the grid energy management system (EMS) as the `presumed topology'. Notably, this topology is not assumed to be perfect and accurate. Topological errors, present in the `presumed topology', are to be detected by the proposed method using power flow measurements from the sensor set $\S$. Note that the proposed methodology is also capable of detecting measurement anomalies, but coordinated measurement anomalies can manifest the same behavior as topology errors, and therefore, classification of the anomaly type (topology vs. measurement) requires additional work. Finally, the proposed methodology enables localization of anomalies, whose accuracy increases with the number of sensors.

\subsection{Context-agnostic Mapping of Power Flow Measurements} \label{subsec:mapping}

The top plot in Figure~\ref{fig:flow_comparison_line89_loadcorrection} shows the raw time-series data of active power flow on a particular line of a European power network. Due to the combined presence of regular (anticipated) topology changes, anticipated load/generation shift events, and (non-anticipated) topology errors, the data signal is irregular and unfavorable for detecting topology errors. 
Here we present a mapping that converts the raw power flow measurement data into \textit{context-agnostic} flow values. The mapping is a three-step process utilizing the linear sensitivity factors (discussed above), and is a key contributor to the superior performance of the detection algorithm proposed in this work.

\vspace{1mm}

\noindent \textbf{Step 0: Selecting the baseline context} 

\noindent The first step is to choose a baseline \emph{context} (injections and topology), which can act as a reference point for the dynamically changing grid context. In essence, the goal is to translate all the measurement data (each coming from its own grid context) into a single unified grid context by using the baseline reference. Let us define $p_0^{\text{inj}} \in \mathbb{R}^{|\V|}$ to be the `baseline injections' and $\mathcal{G}_0=(\V, \E_0)$ to be the `baseline topology.' Note that these baselines are invariant of time and can be chosen arbitrarily but ideally as values with certain central tendencies. The choice of $p_0^{\text{inj}}$ depends on the user, and a potential option would be to use the power flow solution from an average loading scenario. The choice of $\mathcal{G}_0$ again depends on the user, and potential choices would be to choose (i) the graph structure that contains all the edges in the system or (ii) the graph structure that occurs most often. 

\vspace{1mm}

\noindent \textbf{Step 1: Injection-correction of Flow Measurements}

\noindent The second step aims to correct the variability present in the nodal power injection values (which depend on load and generation) so that \textit{known and anticipated} abrupt load/generation changes or shift events are not flagged as anomalies. Let $F(\tau) \in \mathbb{R}^{|\E(\tau) \cap \E_{\S}| \times |\V|}$ be a time-variant matrix (variant due to the dynamic network topology) whose entries contain the PTDF values when the ejection node is fixed as the slack bus. In other words, omitting the dependence on $\tau$, $F(l, i) := f_{l}^{i,\text{slack}}$, where $f_{l}^{i,\text{slack}}$ is the ratio between power change on an observed line $l$ and the power transferred between node $i$ and the slack bus. Also, define $p^{\text{inj}}_\tau \in \mathbb{R}^{|\V|}$ to be the vector of nodal active power injections at time $\tau$. Then, the observed flow values $p_{e,\tau}$ are converted via the following equation: 
\begin{align}
    \hat{p}_{e,\tau} \simeq p_{e,\tau} + F(p_0^{\text{inj}}-p^{\text{inj}}_\tau) \quad \forall e \in \E(\tau) \cap \E_{\S} \label{eqn:PTDF_base}
\end{align}
In essence, $p_{e,0}$ estimates the flow values if the injection values were set to the ‘baseline injections’ while keeping the other factors fixed. Let use denote $\Phi^1(\cdot)$ to be an operator that takes as input $p_{\tau}$, then outputs $\hat{p}_{\tau}$ following equation~\ref{eqn:PTDF_base}. In other words, $\hat{p}_{\tau} = \Phi^1(p_{\tau})$.

\vspace{1mm}

\noindent \textbf{Step 2: Inverse Projection of Flow Measurements} 

\noindent The third step aims to consider the dynamically changing network topology so that regular (anticipated) topology changes are not flagged as anomalies. For simplicity of presentation, we will assume that due to regular (anticipated) topology changes, $\mathcal{G}(\tau)$ contains all nodes and edges of $\mathcal{G}_0$ except for a single edge $k$. Then, the power flows on the edges of $\mathcal{G}(\tau)$ can be approximated by the power flows on the edges of $\mathcal{G}_0$ using the LODF values. 
Now, we only receive measurements from lines in $\E_{\S}$, so the following equations give the system of equations that we are interested in:
\begin{align}
    \hat{p}_{e,\tau} \simeq p_{e,0} + d_e^k p_{k,0} \quad \forall e \in \E(\tau) \cap \E_{\S} \label{eqn:LODF_base}
\end{align}
where $d_e^k$ gives the ratio between power change on an observed line $e$ and the pre-outage real power on the outage line $k$. Note that $\hat{p}_{e,\tau}$ is a fixed parameter obtained from Step 1, whereas $p_{e,0}$ is an unknown variable that we wish to estimate. Therefore, this gives us a linear system of equations with $|\E(\tau) \cap \E_{\S}|$ number of equations and $|\E(\tau) \cap \E_{\S}|+1$ number of variables. By solving this system of equations, we are essentially gauging the edge flow values if the system topology was set to the `baseline topology.' In doing so, we are projecting all the flows associated with the dynamic (i.e., changing) topology to a single fixed `baseline topology,' enabling us to learn a unified statistical model. The remaining problem is that the solution to (\ref{eqn:LODF_base}) is not unique and constitutes a 1-dimensional subspace. 

We will use, $p^{PF}(\mathcal{G}_0, p_0^{\text{inj}})$, the power flow solution from the `baseline topology' with `baseline injections' to overcome this issue. The novelty here is that the solution is independent of the errors or anomalies in the current topology (or the `presumed topology' at the current time) and only depends on the reference \emph{baseline context}. Then, the baseline power flow solution is projected onto the 1-dimensional subspace found in the earlier step. The rationale for doing \textit{this process} is simple: if $\mathcal{G}(\tau)$ contains zero topology errors, the 1-dimensional subspace resulting from solving (\ref{eqn:LODF_base}) will be close to the baseline power flow solution; however, if $\mathcal{G}(\tau)$ contains topology errors, the relationship described in (\ref{eqn:LODF_base}) will be incorrect and produce a 1-dimensional subspace that is far from the baseline power flow solution. 

\begin{remark}
    Step 2 is named inverse-projection because although mechanically, the process is no different from projecting a point onto a 1-D subspace, the core idea embodies the opposite. When the topology change entails two or more edges, the process involves a projection of the point onto a multi-dimensional subspace.
\end{remark}

\begin{remark}
    The user can update the `baseline topology' and `baseline injections' over time when the system's operating point changes significantly, and linearization leads to large inaccuracies (in comparison to the baseline).
\end{remark}



In order to derive a compact form of equation~(\ref{eqn:LODF_base}), first let $\widetilde{\E}(\tau) = \E(\tau) \cap \E_{\S}$, $\widetilde{\E}_0 = \E_0 \cap \E_{\S}$ and define $\hat{p}_{\tau} \in \mathbb{R}^{|\widetilde{\E}(\tau)|}$ to be the vector whose entries are $\hat{p}_{e,\tau}$ for $e \in \widetilde{\E}(\tau)$. Similarly, $x \in \mathbb{R}^{|\widetilde{\E}_0|}$, which takes on the role of $p_0$, is defined to be a vector whose entries collect the power flows over lines that are in the set $\widetilde{\E}_0$. Finally, $d^k \in \mathbb{R}^{|\widetilde{\E}(\tau)|}$ is the vector of LODF values corresponding to the deletion of line $k$. Then, the projection can be performed by solving the following optimization problem.


\begin{equation}
\begin{aligned}
\min_{x \in \mathbb{R}^{|\widetilde{\E}_0|}} & \quad \| x - p^{PF}(\mathcal{G}_0, p_0^{\text{inj}})\| \\
\text{subject to} & \quad \textbf{A}(d^k)x = \hat{p}_{\tau}
\end{aligned}
\tag{P1}
\end{equation}





\let\oldnl\nl
\newcommand{\nonl}{\renewcommand{\nl}{\let\nl\oldnl}}

\begin{algorithm}[!ht]
	\caption{Computation of Graph Distances and Assignment of Weights~\cite{Dynwatch_2021}} 
	\label{alg:weights}
	\KwIn{Time-series graph data $\mathcal{G}_1, \mathcal{G}_2, \ldots, \mathcal{G}_T$ and $\mathcal{T}^A$, the set of time ticks with detected anomalies up until (and excluding) current time tick $\bar{T}$+1. }
	\KwOut{Graph distances $D(\mathcal{G}_1, \mathcal{G}_T)$, $D(\mathcal{G}_2, \mathcal{G}_T)$, \ldots, $D(\mathcal{G}_{T-1}, \mathcal{G}_T)$ and weights $w_1, w_2, \ldots, w_{\bar{T}}$.}
    {\bf Compute the graph distances.}
    
    \nonl \For{$t\leftarrow 1$ \KwTo $T-1$}{
    \nonl \emph{Compute contribution of line $k$ using LODF}\;
    \nonl \For{$k \in (\E(t)-\E(T)) \cup (\E(T)-\E(t))$}{\label{forins}
    \nonl $$x_k = \frac{1}{|\E(t)\cup\E(T)|} \sum_{l\in \E(t)\cup\E(T)\backslash\{k\}}(|d_l^k|)$$
    }
    \nonl \emph{Compute graph distance between $\mathcal{G}_t$ and $\mathcal{G}_T$}\;
    \nonl $$D(\mathcal{G}_t, \mathcal{G}_T) = \sum_{k\in (\E(t)-\E(T)) \cup (\E(T)-\E(t))}x_k$$ 
    }

	{\bf Extend graph distances to tick-wise distances.} Each previous time tick is given a distance $d_{\tau}$ according to the graph it comes from:
	$$d_{\tau} = 
    \begin{cases}
	D(\mathcal{G}_1, \mathcal{G}_T) & \text{for } \tau = 1, \ldots, n_{\tau} \\
	\qquad \vdots & \qquad \qquad \vdots \\
    D(\mathcal{G}_{T-1}, \mathcal{G}_T) & \text{for } \tau = (T-1)n_{\tau}+1 , \ldots, Tn_{\tau} \\
    0 & \text{for } \tau = Tn_{\tau}+1 , \ldots, \bar{T}.
	\end{cases}$$
	
    {\bf Compute optimal lagrange multiplier.} Compute the unique $\lambda^*$ that satisfies:
	$$\sum_{\tau \in\{1,2,\ldots,\bar{T}\}} \max\left\{\frac{\lambda^*-d_{\tau}}{\rho},0\right\}=1$$
	
	{\bf Assign weights $\{w_{\tau}\}$ to the past sensor data.} 
 
    \nonl \For{$\tau \leftarrow 1$ \KwTo $\bar{T}$}{
    \nonl \uIf{$\tau \in \mathcal{T}^A$}{
    \nonl \emph{Set weights corresponding to anomalies to zero}\;
    \nonl $w_{\tau} = 0$
    }
    \nonl \Else{
    \nonl $w_{\tau}= \max\{(\lambda^*-d_{\tau})/\rho,0\}$
    }
    }
    \nonl \emph{Normalize weights such that they sum to one}\;
    \nonl $$w_{\tau} = w_{\tau}/\sum_{\tau} w_{\tau}$$
    \vspace{-1.5mm}
\end{algorithm}

The optimal solution, $x^*$, represents the best candidate for $p_0$ in the sense that it is the closest we can get to the baseline powerflow solution, while satisfying the LODF equations. In the above problem, $\textbf{A}(d^k)$ is a matrix that is structured like an identity matrix separated in the middle by the column vector $d^k$, and is designed to encode equation~(\ref{eqn:LODF_base}).
Note that in the case that we consider prevalently in this paper, $|\widetilde{\E}_0|$ = $|\widetilde{\E}(\tau)|$+1, and therefore the dimension of Null($\textbf{A}(d^k)$) is equal to one. Let us denote $\Phi^2(p^{PF}(\mathcal{G}_0, p_0^{\text{inj}}), \hat{p}_{\tau})$ to be an operator that takes as input $p^{PF}(\mathcal{G}_0, p_0^{\text{inj}})$ and $\hat{p}_{\tau}$, then outputs $\check{p}_{\tau}$, the optimal solution of (P1). In other words, $\check{p}_{\tau} := x^* = \Phi^2(p^{PF}(\mathcal{G}_0, p_0^{\text{inj}}), \hat{p}_{\tau})$.
We provide a detailed discussion on the impact of these mappings in later sections, but for a quick glimpse of the results, one can compare the line flow on branch 89 after correction in Step 2 in the middle subplot of Figure~\ref{fig:flow_comparison_line89_loadcorrection} and line flow in branch 89 after corrections in Steps 1 and 2 in the bottom subplot of Figure~\ref{fig:flow_comparison_line89_loadcorrection}. The top subplot shows the uncorrected observed line flow on branch 89.

\subsection{Algorithm}
Given measurement data at time $\bar{T}+1$ (current time), the goal is to use historical time-series data from time ticks $\tau = 1, \ldots, \bar{T}$ and detect any topological or measurement anomalies present in the SCADA data. We also assume that the power grid undergoes regular (anticipated) topology changes implemented at the beginning of each period. Each period is $n_{\tau}$ time ticks long, and the current time tick is in the $T$-th period. 

The overall anomaly detection consists of two parts; The first part (Algorithm~\ref{alg:weights}) involves computing the graph distances and assigning the corresponding weights to each historical data point~\cite{Dynwatch_2021}. Since topology only changes every period, the graph distance needs to be computed for only $T-1$ instances. The weighting scheme in steps 3 and 4 of Algorithm~\ref{alg:weights} is equivalent to solving the following bias-variance tradeoff problem, where $\rho$ denotes the weight put on the variance part. 
\begin{equation}
\begin{aligned}
\min_{w} & \quad w^Td + \rho \|w\|_2^2 \\
\text{subject to} & \quad w^T \mathbf{1} = 1 \\
& \quad w \geq \mathbf{0}
\end{aligned}
\end{equation}
The second part (Algorithm~\ref{alg:detection}) involves applying the context-agnostic mapping introduced in Section~\ref{subsec:mapping} to the time-series data, weighting the historical data to learn a statistical model, and finally detecting any anomalies that are present. The value of the threshold, $\bar{\zeta}$, can significantly impact the algorithm's performance. In practice, one can choose the optimal value of the threshold using cross-validation~\cite{Thresh_2014}.

\begin{algorithm}[!ht]
	\caption{Anomaly Detection via Context-agnostic Mapping} 
	\label{alg:detection}
	\KwIn{Current sensor measurements; past sensor measurements and their weights $w_1, w_2, \ldots, w_{\bar{T}}$ obtained from Algorithm~\ref{alg:weights}.}
	\KwOut{Topological anomalies at time $\bar{T}+1$.}
    {\bf Compute inverse-projection of past measurements.}

    \nonl \For{$t\leftarrow 1$ \KwTo $T$}{
    \nonl Run power flow (PF) on baseline topology and injections and store the relevant flow values in $p^{PF}(\mathcal{G}_0)$\;
    \nonl \uIf{$t \leq T-1$}{
    \nonl \For{$\tau \leftarrow (t-1)n_{\tau}+1$ \KwTo $tn_{\tau}$}{
    \nonl $$\hat{p}_{\tau} = \Phi^1(p_{\tau}), \ \check{p}_{\tau} = \Phi^2(p^{PF}(\mathcal{G}_0), \hat{p}_{\tau})$$
    }
    }
    \nonl \Else{
    \nonl \For{$\tau \leftarrow Tn_{\tau}+1$ \KwTo $\bar{T}$}{
    \nonl $$\hat{p}_{\tau} = \Phi^1(p_{\tau}), \ \check{p}_{\tau} = \Phi^2(p^{PF}(\mathcal{G}_0), \hat{p}_{\tau})$$
    }
    }
    }

    {\bf Create distribution for each measurement sensor.}

    \nonl \For{$i\leftarrow 1$ \KwTo $|\widetilde{\E}_0|$}{
    \nonl $$\mu_i = \text{Weighted mean} \{p_{\tau, i} \ | \ \tau =1,\ldots, \bar{T}\},$$
    \nonl $$\sigma_i = \text{Weighted stdev} \{p_{\tau, i} \ | \ \tau =1,\ldots, \bar{T}\},$$
    where the weights $\{w_{\tau}\}$ obtained from Algorithm~\ref{alg:weights} are used.
    }
    
    {\bf Compute anomaly score for current time tick.}
    Compute context-agnostic ouput of current measurements\;
    \vspace{-4mm}
    \nonl $$\hat{p}_{\bar{T}+1} = \Phi^1(p_{\bar{T}+1}), \ \check{p}_{\bar{T}+1} = \Phi^2(p^{PF}(\mathcal{G}_0), \hat{p}_{\bar{T}+1})$$
    Compute aggregate anomaly score\;
    \nonl $$\zeta = \max_{i} \left|\frac{\check{p}_{\bar{T}+1, i} - \mu_i}{\sigma_i} \right|$$ 
    
    {\bf Detect anomalies.}
    
    \nonl \uIf{$\zeta > \bar{\zeta}$}{
    \nonl Classify current time tick as anomalous\;
    \nonl $$\mathcal{T}^A \leftarrow \mathcal{T}^A \cup \{\bar{T}+1\}$$
    }
    \vspace{0mm}
\end{algorithm}

\section{Numerical Analysis and Discussion}
\subsection{Load Data Generation} \label{subsec:load_data}
In this paper, we use the same load data that was used in \cite{hooi2018gridwatch, Dynwatch_2021}. This dataset consists of a 5s interval time series of loads (i.e., real and reactive power at each node) generated using the patterns estimated from two real datasets listed below.
\begin{itemize}
    \item 24-hour dataset of AC state estimation output data from a real utility in the Eastern Interconnect of the U.S. The dataset contains all operational data of the grid based on a 30-minute interval. See Section V-A of~\cite{Dynwatch_2021} for its dataset description and statistical findings.
    \item Carnegie Mellon University (CMU) campus load data recorded for $20$ days from July 29 to August 17, 2016.
\end{itemize}
In creating the synthetic load profiles, the overall periodic pattern is extracted from the utility-provided dataset, while the distribution of noise is extracted from the CMU dataset. The final load data is generated by choosing a specific testcase, scaling the magnitude of daily load variation to a predefined level, and adding Gaussian noise sampled from the extracted noise distribution~\cite{song2017powercast}. The testcase that we use extensively throughout the rest of the paper is the Polish power grid (often referred to \textbf{\emph{case2383wp}}), which contains 2383 number of buses and 2896 number of lines. 

\vspace{-1.5mm}

\subsection{Experimental Settings} 
Starting with a particular test case as a baseline graph $\mathcal{G}_0$, we first create $20$ topology scenarios representing the dynamically changing network, where each of them deactivates a randomly chosen branch in the baseline graph. This sequence of $20$ network topologies represents the topology changes resulting from regular operation and control of the system operator. Then, for each topology scenario, we use MATPOWER~\cite{zimmerman2011matpower} to generate $60$ sets of synthetic measurements based on the load characteristics described in Section~\ref{subsec:load_data}. As a result, the multivariate time series data with 1200 ($20\times60$) time ticks mirrors the real-world setting where sensors receive measurements at each time tick $\tau$ (e.g., every 5-15 seconds \cite{Dynwatch_2021}), and the anticipated or known grid topology changes occur $60$ time ticks (e.g., every 5-15 min). Finally, we randomly sample $20$ ticks out of the $1200$ as times when anomalies occur. Each of these anomalies is modeled by randomly deleting an edge on the corresponding topology.


\begin{figure}[t]
	\centering
\includegraphics[width=0.9\linewidth]{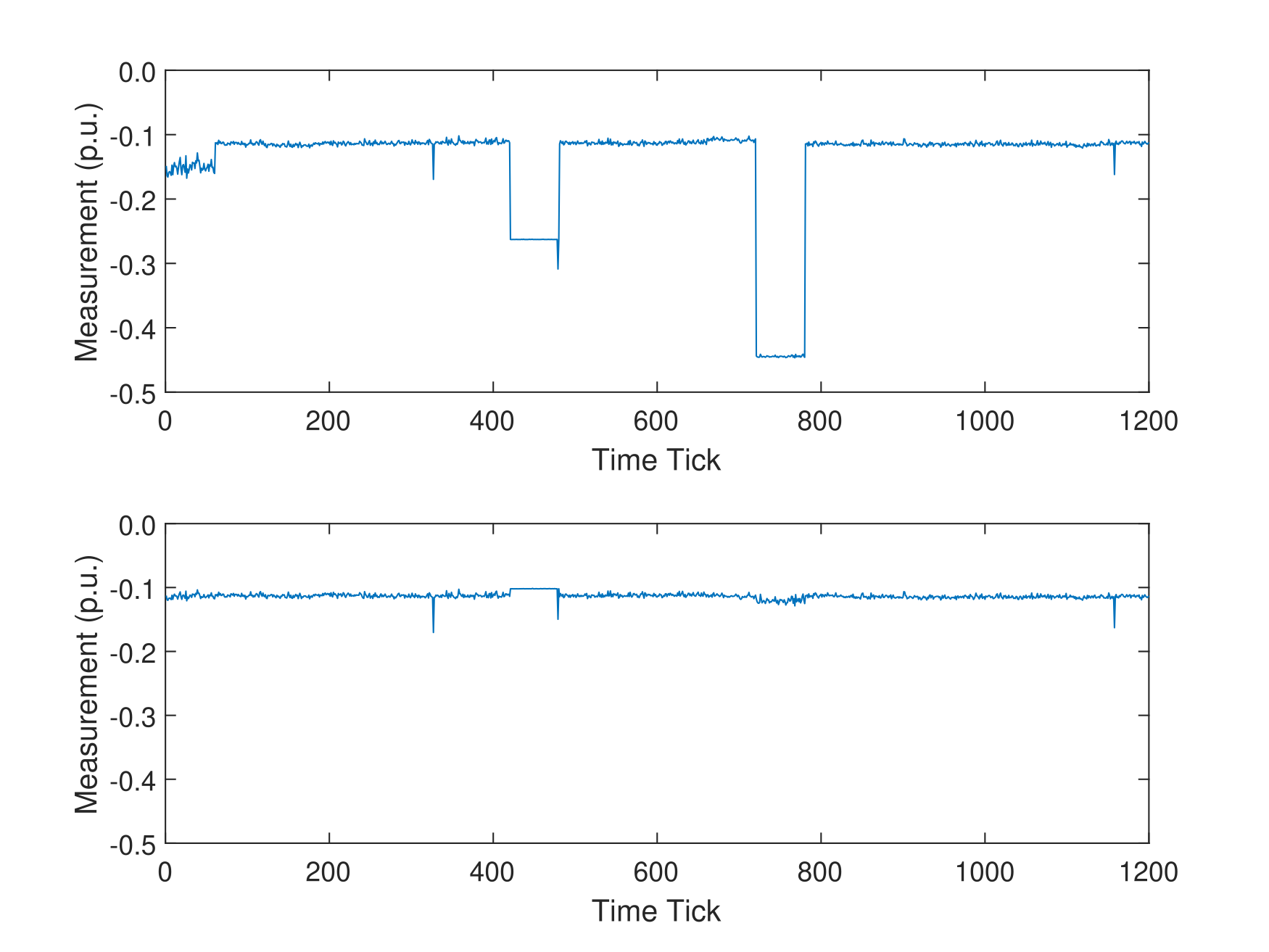}
	\caption[]{Flow signal before (top) and after (bottom) inverse projection on line 24 (\#1 among current edgeset)}
	\label{fig:flow_comparison_line24}
\end{figure}


\subsection{Output of Inverse-projection} \label{subsec:output_inv_proj}
The injection-correction mapping is performed before the inverse projection, but for better delivery, we discuss the latter first.
The efficacy of the LODF-based inverse projection (see Section~\ref{subsec:mapping}) is a critical contributing factor to the overall performance of the anomaly detection algorithm. Here, we showcase the efficacy of the proposed method. As mentioned earlier, all numerical results are performed on the \textbf{\emph{case2383wp}} Polish grid, unless stated otherwise. 



Figure~\ref{fig:flow_comparison_line24} shows an example of a line whose measurement value is affected by one regular topology change and several anomalies.  The anomaly at time 327 (first spike in the figure) is a topology error (unanticipated and anomalous) at line 882, and the anomaly at time 479 (second spike in the figure) is a topology error (anticipated and known) at line 22. In the top plot, it is difficult to differentiate the regular topology changes from the anomalies. The bottom plot, which shows the inverse-projected measurement values, highlights a drastic change; the deviations are accounted for and do not result in a false detection. In a way, the inverse-projection has offset the effect of the regular topology change. 


\subsection{Effect of Injection-correction} \label{subsec:effect_load_corr}

Injection projection can offset the effect of known or anticipated topology changes but cannot account for known and anticipated load/generation shifts. The injection-correction mapping can offset (i) anticipated load/generation shift events and (ii) normal load/generation variability. 

Figure~\ref{fig:flow_comparison_line89_loadcorrection} shows the time-series flow signal of a line before processing (top), after only inverse projection (middle), after both injection-correction and inverse projection (bottom), in the presence of anticipated topology changes, anomalous (non-anticipated) topology changes, and a single load/generation shift event. In the middle plot, we can observe that the inverse projection can offset the anticipated topology changes but cannot offset the anticipated load/generation shift events (at time tick 400) and will, therefore, incorrectly classify the peak at time 400 as an anomaly. On the contrary, the bottom plot shows that with the additional injection-correction step, the peak at time 400 is offset, and we can correctly use the remaining three peaks to detect true topological anomalies. 

\begin{figure}[t]
	\centering
	\includegraphics[width=0.9\linewidth, trim={0 0.7cm 0 0},clip]{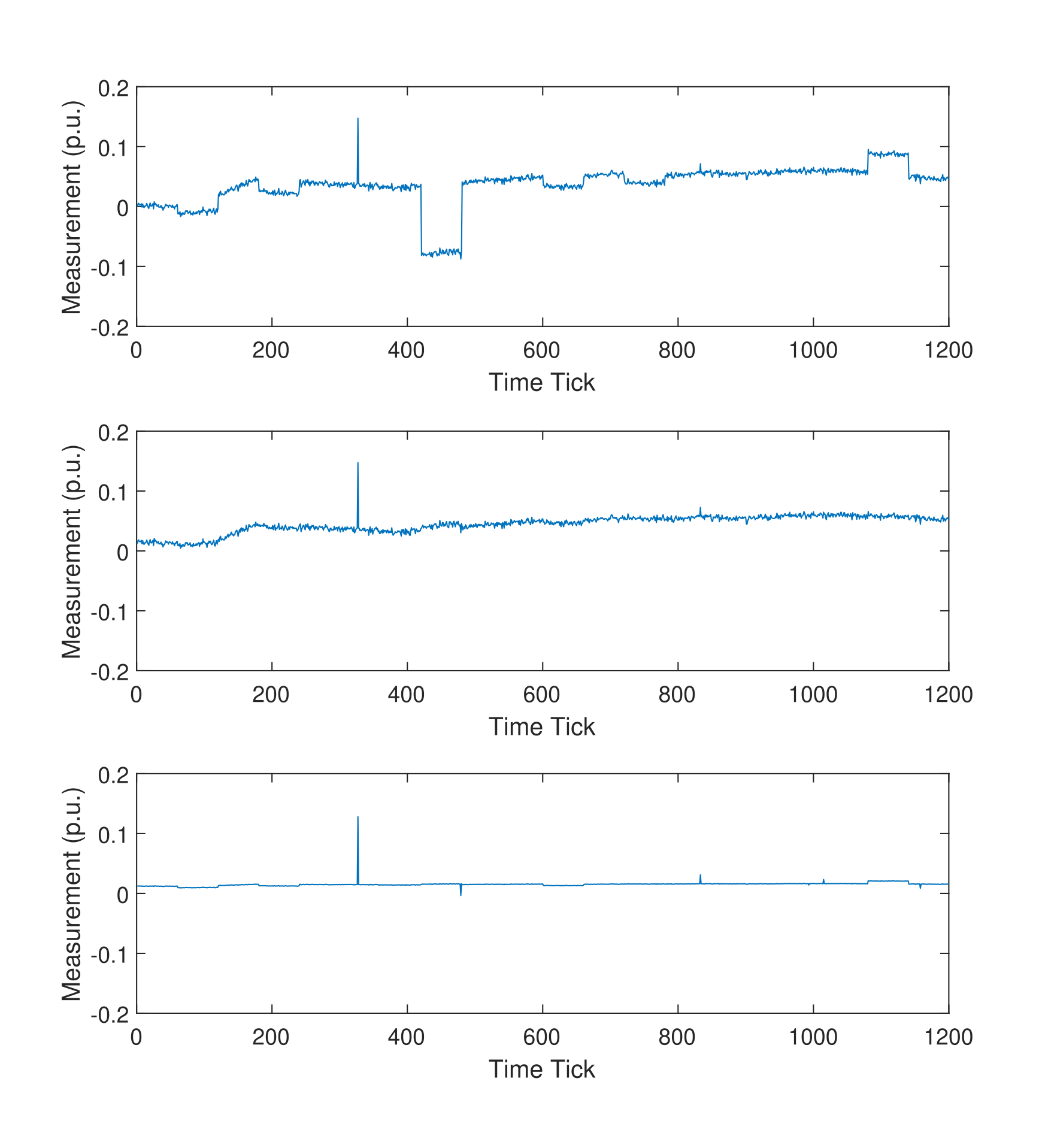}
	\caption[]{Flow signal before any processing (top),  after only inverse projection (middle), and after injection-correction and inverse projection (bottom) on line 93 (\#14 among current edgeset)}
	\label{fig:flow_comparison_line93_loadcorrection}
\end{figure}

Figure~\ref{fig:flow_comparison_line93_loadcorrection} shows similar plots but now in the presence of anticipated topology changes, anomalous (unanticipated) topology changes, and higher load variability. By comparing the bottom plot with the two other plots, one can observe that the variability in load is offset. Without the injection-correction step, steep load changes (such as during time ticks 100 $\sim$ 200) can also be incorrectly identified as anomalies.

\subsection{Comparison with Existing Methods} \label{subsec:existing_methods}
Dynamic graph anomaly detection approaches suggested in \cite{akoglu2010oddball,chen2012community,araujo2014com2,shah2015timecrunch} cannot be used for comparison since they consider graph structure only, but not sensor data. The work \cite{mongiovi2013netspot} utilizes sensor data but requires graphs with fully observed edge weights, which is inapplicable as detecting failed power lines with all sensors present simplifies the problem of checking if any edge has a current equal to $0$. Hence, instead, we compare our proposed method to GridWatch~\cite{hooi2018gridwatch}, an anomaly detection approach for sensors on a static graph, DynWatch~\cite{Dynwatch_2021}, a pioneering work for sensors on a dynamic graph (also the predecessor of the current work), and the following multidimensional time-series based anomaly detection methods: Vector Autoregression (VAR)~\cite{hamilton1994time}, Local Outlier Factor (LOF)~\cite{breunig2000lof}, and Parzen Window~\cite{parzen1962estimation}. Each uses the currents and voltages at the selected sensors as features. For VAR, the norms of the residuals are used as anomaly scores; the remaining methods return anomaly scores directly.
For VAR, adhering to standard practice, we select the order by maximizing the Akaike information criterion (AIC). For LOF, we use $20$ neighbors (following the default in scikit-learn), and we use $20$ neighbors for Parzen Window. GridCAL-naive represents the algorithm presented in this paper, but without the injection-correction and inverse projection, GridCAL-IP is the algorithm with only the inverse projection, and GridCAL-IPLC is with both inverse projection and injection-correction.

To set up the comparative analysis, we randomly sample 20 time ticks out of the entire 1200, and also randomly sample the corresponding anomalous lines. For each approach, the anomaly detection algorithm is run multiple times while varying the number of available sensors. 
Each algorithm returns a ranking of the anomalies. We evaluate this using standard metrics, AUC (area under the ROC curve) and F-measure ($\frac{2 \cdot \text{precision} \cdot \text{recall}}{\text{precision+recall}}$),
the latter computed on the top 20 anomalies output by each
algorithm. 

\begin{figure}[t]
	\centering
	\includegraphics[width=0.85\linewidth, trim={0 0.2cm 0 0},clip]{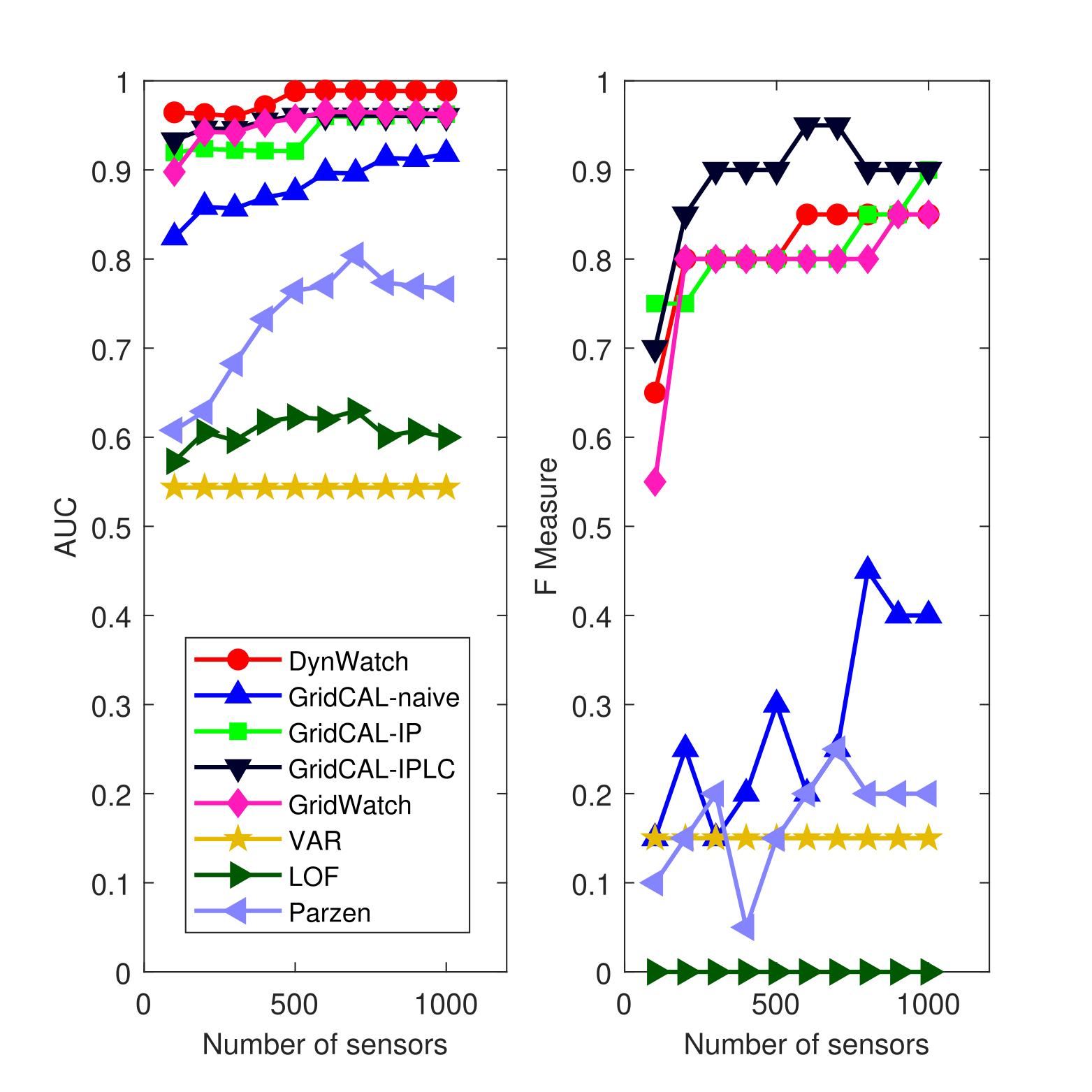}
	\caption[]{Comparing performance of the proposed method with existing methods under the measure of AUC (left) and F-measure (right). The simulations were performed for a dataset with relatively small load variability and no load/generation shift scenarios.}
	\label{fig:comparison1}
\end{figure}

\begin{figure}[t]
	\centering
	\includegraphics[width=0.85\linewidth, trim={0 0.2cm 0 0},clip]{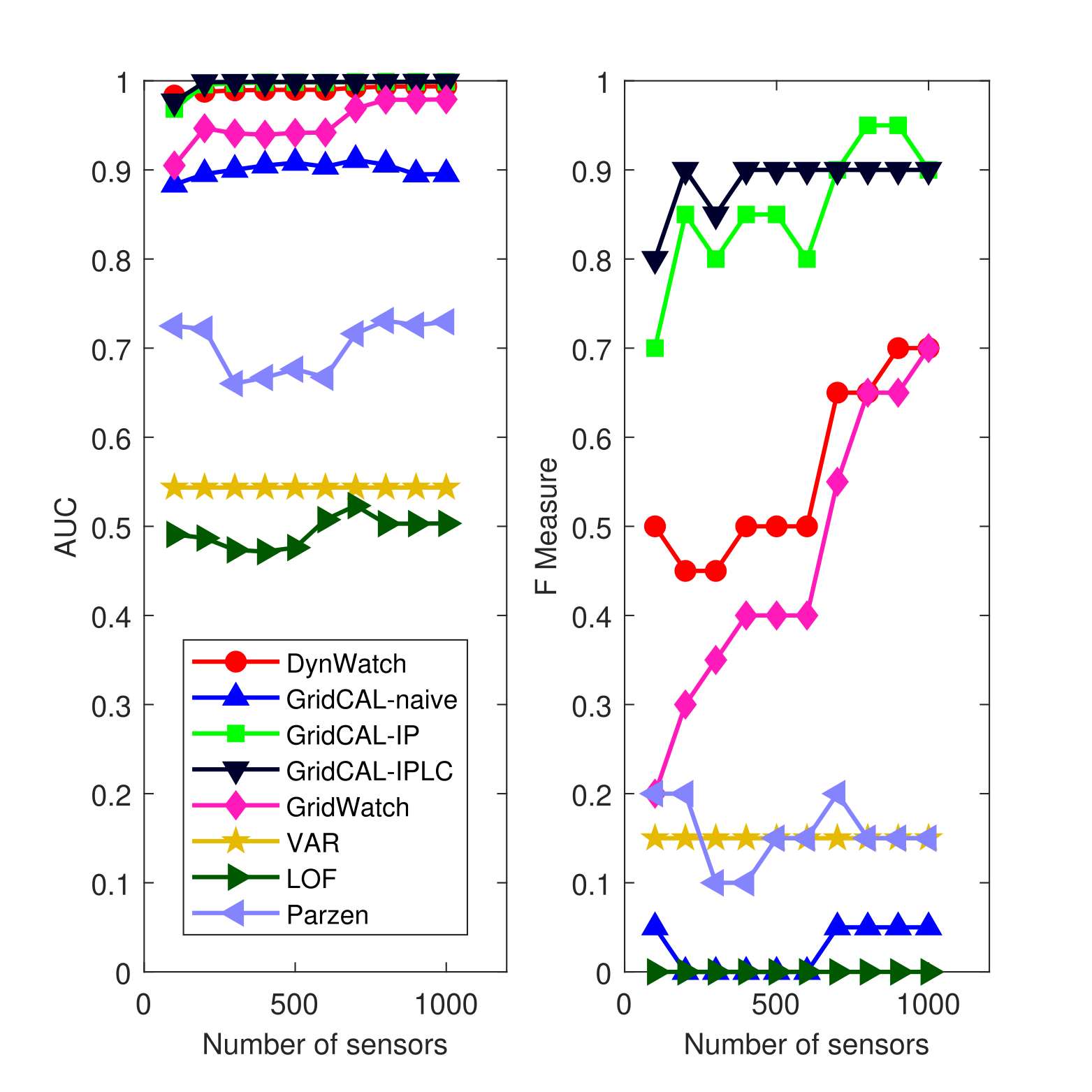}
	\caption[]{Comparing performance of the proposed method with existing methods under the measure of AUC (left) and F-measure (right). The simulations were performed for a dataset with relatively large load variability but without load/generation shift scenarios.}
	\label{fig:comparison2}
\end{figure}

Figure~\ref{fig:comparison1} shows the AUC and F-measure for the eight different methods that are considered. The scenario considered has low load variability and no load/generation shift events. In terms of the AUC, we observe that DynWatch, GridCAL-IP, GridCAL-IPLC and GridWatch work significantly better compared to the other approaches. With regards to the F-measure, GridCAL-IPLC works significantly better than others, while DynWatch, GridCAL-IP, and GridWatch trail very closely. The impact that the number of sensors has on the performance is highlighted in the F-measure curves. We observe that the F-measure generally trends higher with an increasing number of sensors. However, the curve for GridCAL-IPLC is strictly above that of the other methods, which displays the superiority of the proposed method.  

\begin{figure}[t]
	\centering
	\includegraphics[width=0.85\linewidth, trim={0 0.2cm 0 0},clip]{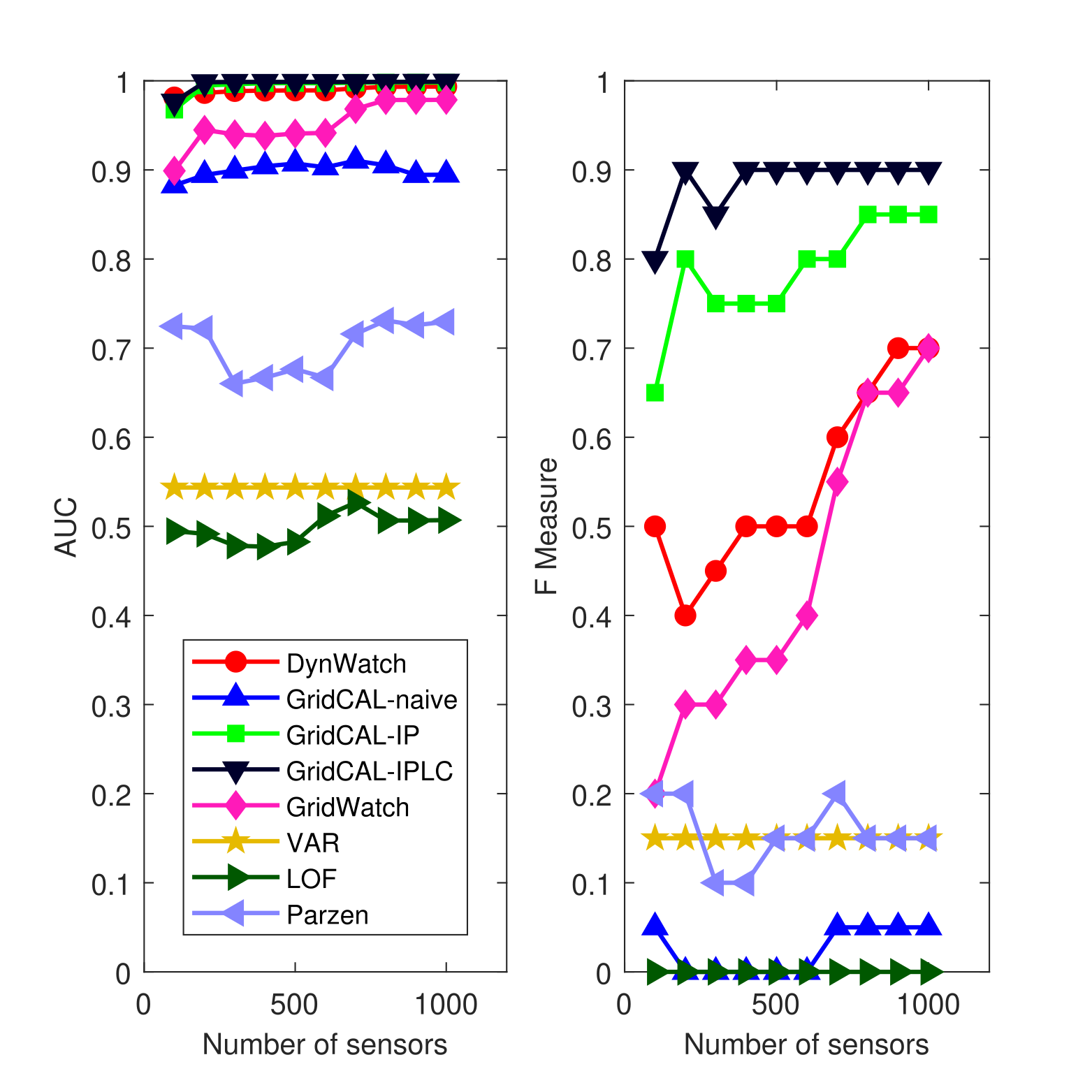}
	\caption[]{Comparing performance of the proposed method with existing methods under the measure of AUC (left) and F-measure (right). The simulations were performed for a dataset with relatively large load variability and load/generation shift scenarios.}
	\label{fig:comparison3}
\end{figure}

Figure~\ref{fig:comparison2} shows the same plots but under the scenario where there is high load variability but still no load/generation shift events. In this case, we can observe from the F-measure that the performance of DynWatch and GridWatch deteriorates, whereas GridCAL-IP and GridCAL-IPLC are still performing very well. This is accounted for by the fact that DynWatch and GridWatch have high false-positive rates (Type 1 error). Finally, Figure~\ref{fig:comparison3} shows the same plots but under the scenario where there is high load variability and a number of (10) load/generation shift events. In this case, we can observe from the F-measure that the performance of GridCAL-IP falls behind that of GridCAL-IPLC. As both anticipated load/generation shifts (e.g., generator trip or re-dispatch, sudden load increase) and topology shifts (e.g., transmission line trip or switching) are common occurrences, GridCAL-IPLC is likely to work best in practical, real-world scenarios.

\section{Conclusion} \label{sec:conclusion}
This paper presents a \textit{context-aware} anomaly detection algorithm based on unsupervised learning. We propose a two-step mapping that converts the real-time power flow measurement data into \textit{context-agnostic} flow values, allowing us to analyze data from the dynamic context in an aggregate fashion. The methodology performs superior to existing methods, especially in the presence of anticipated topology changes, anticipated load/generation shift events, and high load/generation variability.

\bibliographystyle{IEEEtran}
\bibliography{references}

\end{document}